# The Conquest of Quantum Genetic Algorithms: The Adventure to Cross the Valley of Death

Rafael Lahoz-Beltra[1,2]

[1]Department of Biodiversity, Ecology and Evolution (Biomathematics), Faculty of Biological Sciences, Complutense University of Madrid, 28040 Madrid, Spain

[2]Modeling, Data Analysis and Computational Tools for Biology Research Group, Complutense University of Madrid, 28040 Madrid, Spain

[*]Address correspondence to: lahozraf@ucm.es

**Abstract**

In recent years, the emergence of the first quantum computers at a time when AI is undergoing a fruitful era has led many AI researchers to be tempted into adapting their algorithms to run on a quantum computer. However, in many cases the initial enthusiasm has ended in frustration, since the features and principles underlying quantum computing are very different from traditional computers. In this paper, we present a discussion of the difficulties arising when designing a quantum version of an evolutionary algorithm based on Darwin's evolutionary mechanism, the so-called genetic algorithms. The paper includes the code in both Python and QISKIT of the quantum version of one of these evolutionary algorithms allowing the reader to experience the setbacks arising when translating a classical algorithm to its quantum version. The algorithm studied in this paper, termed RQGA (Reduced Quantum Genetic Algorithm), has been chosen as an example that clearly shows these difficulties, which are common to other AI algorithms.

1.  Introduction

In recent years, the concurrence of the first quantum computers and the current advances in AI has led to many researchers to attempt the migration of these AI algorithms to a quantum computer. Since their early days, AI algorithms have been inspired by Nature, using principles both from human and animal learning and from the evolution of living beings. However, many of these bio-inspired principles that have been successfully emulated in classical computers, i.e. digital computers, are facing serious difficulties [1] when we try to deploy their counterparts in a quantum computer.

During the 1940s and 1950s the concepts of intelligence and evolution were used interchangeably by scientists such as John von Neumann or Alan Turing. Over time, disciplines such as Artificial Intelligence, or later Artificial Life, adopted such concepts in a similar way as these scientists did after World War II. Moreover, today, the procedure adopted to deal with multidisciplinary problems with machine learning techniques still retains the approach adopted by Turing and von Neumann. From a practical point of view, this way of understanding what intelligence is allowed to Alan Turing [2] introduce his famous test, the Turing test, adopting a behaviorist approach. Consequently, by adopting a behaviorist approach, Turing avoided having to define what intelligence is contributing in an unpremeditated way to an indistinct use of the concepts of intelligence and evolution. Even today, AI algorithms are applied based on the old understanding of the concepts of intelligence and evolution and therefore from a purely instrumental perspective. Therefore, intelligence is understood as the ability to learn and solve specific tasks such as mathematical problems, translating a text from one language to another, playing chess, etc. Likewise, and under the IA approach, learning is considered to consist of given a certain input, e.g. a signal, the ability of the system to change the output, e.g. a response or behavior. This point of view is the one adopted by artificial neural network models (Figure 1), in which learning is simulated by adjusting the connections between the nodes that model the neurons by applying a certain algorithm for this purpose [3]. In addition, and as far as evolution is concerned, nowadays we have optimization algorithms that under the common name of evolutionary algorithms are successfully applied in combinatorial and optimization problems of different disciplines. Normally, evolution is simulated by drawing inspiration from the mechanism of natural selection proposed by Darwin. One of these algorithms, perhaps one of the most popular, are the so-called genetic algorithms [3, 4]. Unlike neural networks, while in an artificial neural network the information that is modified during learning is stored in the weights of the connections, in a genetic algorithm (Table I) the information to be selected during the emulation of the evolutionary process is stored in vectors representing the chromosomes (Figure 2) of an organism. Throughout the simulation time, the vectors will be selected for their suitability to be part of the next generation, and the information contained will be transformed by operators that simulate genetic mechanisms, such as mutation, recombination etc.

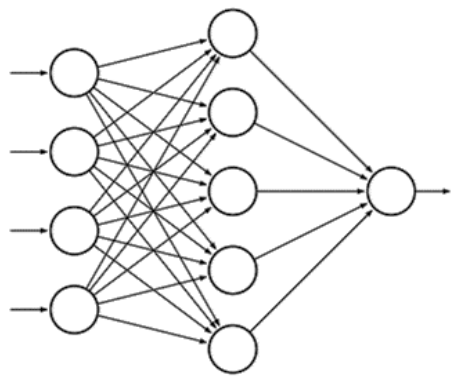

**Figure 1**. Artificial neural network depicting an input layer with four neurons that receive the input signal, and an intermediate or hidden layer with five neurons all connected to the input layer. The network includes an output layer formed by a single neuron responsible for the output or response after information processing by the neural network.

To date, genetic algorithms (GA) have been implemented in a multitude of programming languages and have been widely applied to solve the most diverse real-world problems. However, the success of artificial neural networks and genetic algorithms is due to the fact that they are executed on the computers we use today. Today's computers are digital machines that are the result of a rapid evolution since the appearance of the first computers in the 1950s. The computer as it is still conceived today has allowed the implementation and execution of the most diverse algorithms under a model known as von Neumann architecture (Figure 3).

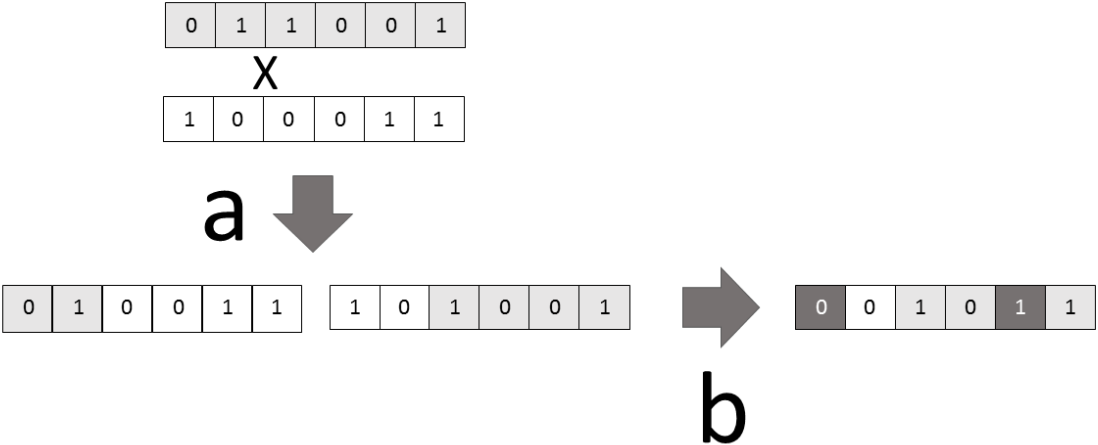

**Figure 2**. (a) One-point recombination. Once two parental chromosomes have been chosen, a point where the genes intersect—for example, at the second position—is randomly selected. The chromosome segments are then exchanged, resulting in two recombinant chromosomes, i.e., the offspring. Note how the offspring chromosomes share the information of the parental chromosomes. (b) Mutation. In one of the child chromosomes, specifically the right one, we choose two positions at random, the first and the fifth in the example, inverting the value of the binary number.

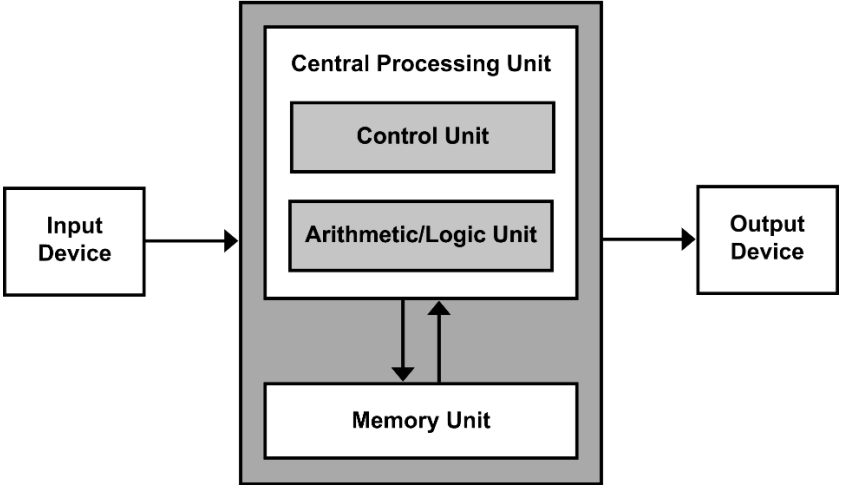

**Figure 3**. von Neumann architecture showing the different components of a microprocessor (taken from en.wikipedia.org )

In 2018 began the so-called NISQ era of quantum computing [5] in reference to the fact that the quantum computers either in existence or in the pipeline at the time were of intermediate size and noisy. In this context the term noise referred to the fact that the computers displayed a still imperfect control of the qubits. Of the three available quantum computers, i.e. D-wave [6], Google [7] and IBM [8], we will focus the discussion presented in this article on the last of them, that is, on the IBM quantum computer. The advent of these computers initially led to the belief that machine learning and optimization algorithms could be quickly upgraded to their quantum version in order to take advantage of the benefits of this new class of computers. At last, and from now on, AI algorithms would run faster and with lower power consumption. Although these quantum computer qualities are indeed true, Can an algorithm designed to run on a classical computer be translated into its counterpart for a quantum computer? What difficulties arise? Can we translate command by command, subroutine by subroutine, one by one each of the steps of the algorithm?

In this paper we will address one of the most interesting challenges resulting from the recent appearance and availability of the earliest quantum computers. Currently, some researchers involved in the application of AI algorithms are trying to "translate" these algorithms into their equivalents to be executed on a quantum computer. Artificial neural network models [9] and genetic algorithms are two clear examples that illustrate this effort, since this would make it possible to solve classification and optimization problems on a quantum computer by taking advantage of the remarkable benefits of quantum computing. In this article, and through the author's personal experience, we will present the pitfalls we are currently facing in designing the quantum version of a genetic algorithm. The aim of the present article is to present the current state of the field in order to show how the "marriage between AI and quantum computers" is not an easy road to take, at least with the current state of the art. First, we will present what a classical genetic algorithm (GA) is. Second, we will expose how to implement the quantum GA (QGA) version under the QRAM quantum computing paradigm (Figure 4). Next, we will outline the standard version of a QGA to be executed on a computer with von Neumann architecture, i.e. on a classical non-quantum computer. In third place, we will discuss the main difficulties encountered when running the standard version of the QGA on a quantum computer. Finally, we will conclude this discussion by experimenting with the RQGA quantum algorithm on both a classical computer and the IBM quantum computer. RQGA is to date an interesting proposal of a quantum genetic algorithm that was suggested years ago with the aim to overcome the difficulties arising when a QGA is run on a quantum computer. In the article we include an Appendix with the codes in Python and Qiskit [10] languages so the interested reader can experience with a QGA and RQGA respectively.

## 1.1. Genetic algorithms: classical vs. quantum versions

Genetic algorithms (GA) are optimization methods inspired by Darwin's theory of evolution [3, 4], which are used in optimization problems, such as the design of objects, devices or machines in which one or more of their characteristics must be optimized. For example, the design of an antenna, the shape of the body of a car, the shape of a molecule, an electronic circuit, etc. Although genetic algorithms are already a well-known optimization technique, it is still possible to innovate by making original proposals. For example, [11] proposed a GA inspired by the genetic mechanisms of bacteria designing an AM radio receiver by simulating the bacterial conjugation mechanism.

**Table I.** Genetic algorithm

| Step | |
|---|---|
| 1 | Randomly initialize a population P(0) |
| 2 | Evaluate P(0) |
| 3 | while (not termination condition) do |
| 4 |   begin |
| 5 |     t ← t + 1 |
| 6 |     Selection of parents from population P(t) |
| 7 |     Crossover |
| 8 |     Mutation |
| 9 |     Evaluate P(t) |
| 10 |   end |

A simple genetic algorithm (SGA) comprises the following steps (Table I). First, (a) a population of solutions is randomly generated. The solutions are vectors representing chromosomes, sub-strings representing genes. These vectors can hold 0s and 1s, integers, real numbers, etc. Next, (b) the goodness or validity of the solutions is evaluated. This goodness, called fitness, represents in Nature a measure of the adaptation of an individual to the environment. In an optimization problem the environment is the problem space. Solutions with a higher fitness value will have a higher probability of surviving and, consequently, (c) of reproducing. Consequently, the good solutions will have a higher chance of being selected, reproduce and pass to the next generation, propagating the good solutions to the offspring through the genes ("substrings") of the chromosomes. However, in order to have variability without which evolution cannot succeed, the individuals, chromosomes or solutions that pass to the next generation (d) will be modified by genetic mechanisms (Figure 2). These mechanisms are mainly two: crossover and/or mutation giving. Therefore, the offspring will be genetically different from the parental generation from which it comes. The search process shown in Table I is repeated over a number of generations and ends when the termination criterion is reached. For example, the search ends when the solution is found, i.e. when the optimal or near-optimal chromosome is found or when a given maximum number of generations is reached.

**Table II.** Quantum genetic algorithm

| Step | Quantum Computing | Classical Computing |
|---|---|---|
| 1 | Initialize a quantum population Q(0) | |
| 2 | Make P(0), measure of every individual Q(0) → P(0) | |
| 3 | | Evaluate P(0) |
| 4 | while (not termination condition) do | |
| 5 |   begin | |
| 6 |     t ← t + 1 | |
| 7 |     Rotation Q-gate | |
| 8 |     Mutation Q-gate | |
| 9 |     Make a measure Q(t) → P(t) | |
| 10 | | Evaluate P(t) |
| 11 |   end | |

The first quantum genetic algorithm (QGA) was proposed in 2002 by [12] which encouraged the research and design of an SGA quantum version. We will assume that the concepts and principles on which quantum computing is based are familiar to the reader at an elementary level. If so, we will disclose below the main steps of a quantum genetic algorithm (QGA). The plain version of a QGA comprises the following steps (Table II):

The first step involves the initialization of a quantum population $Q(0)$ of chromosomes. A quantum chromosome $i$ is defined as a string of $j$ qubits representing a quantum system with $2^j$ simultaneous states:

$$|\psi\rangle^i = \begin{pmatrix} \alpha_1 & \alpha_2 & \dots & \alpha_j \\ \beta_1 & \beta_2 & \dots & \beta_j \end{pmatrix} = \sum_j c_i |\psi_j\rangle^i \quad (1)$$

where $|\psi_j\rangle^i$ represents de $j$ gene in $i$ chromosome:

$$|\psi_j\rangle^i = \begin{pmatrix} \alpha_j \\ \beta_j \end{pmatrix} \quad (2)$$

The most common technique to initialize the population is to set the value of the amplitudes of all qubits in the chromosomes to a value representing the quantum superposition of all states with equal probability. This is achieved from the product of the Hadamard matrix by the vector $|0\rangle$, obtaining the following superposition vector:

$$H|0\rangle = \frac{1}{\sqrt{2}} \begin{pmatrix} 1 & 1 \\ 1 & -1 \end{pmatrix} \begin{pmatrix} 1 \\ 0 \end{pmatrix} = \frac{1}{\sqrt{2}} \begin{pmatrix} 1 \\ 1 \end{pmatrix} \quad (3)$$

Subsequently to this product a phase angle is randomly obtained. Phase angle is the argument of the trigonometric functions or elements in the rotation matrix $U(t)$:

$$|\psi_j\rangle^i = \begin{pmatrix} \alpha_j \\ \beta_j \end{pmatrix} = \begin{pmatrix} Cos(\theta_j) & -Sin(\theta_j) \\ Sin(\theta_j) & Cos(\theta_j) \end{pmatrix} \frac{1}{\sqrt{2}} \begin{pmatrix} 1 \\ 1 \end{pmatrix} \quad (4)$$

In a second step we obtain a classical GA population $P(t)$, thus a population of vectors (Table III) as is usual in a simple or standard genetic algorithm (SGA). The population $P(t)$ is the result of the measure or observation of qubits states in the quantum chromosomes of the population $Q(t)$. In consequence, after measurement we obtain the classical population $P(t)$ which is given by a set of $i$ vectors $(x_1, x_2, \dots, x_j)$:

**Table III**. Obtaining *P*(t) after observing or measuring *Q*(t)

| $Q(t)$ | $\rightarrow$ | $P(t)$ |
|---|---|---|
| $\begin{pmatrix} \alpha_1 & \alpha_2 & \alpha_3 & \ldots & \alpha_j \\ \beta_1 & \beta_2 & \beta_3 & \ldots & \beta_j \end{pmatrix}_1$ | | $(x_1\ x_2\ x_3 \ldots x_j)_1$ |
| $\begin{pmatrix} \alpha_1 & \alpha_2 & \alpha_3 & \ldots & \alpha_j \\ \beta_1 & \beta_2 & \beta_3 & \ldots & \beta_j \end{pmatrix}_2$ | | $(x_1\ x_2\ x_3 \ldots x_j)_2$ |
| , ... , | | , ... , |
| $\begin{pmatrix} \alpha_1 & \alpha_2 & \alpha_3 & \ldots & \alpha_j \\ \beta_1 & \beta_2 & \beta_3 & \ldots & \beta_j \end{pmatrix}_i$ | | $(x_1\ x_2\ x_3 \ldots x_j)_i$ |

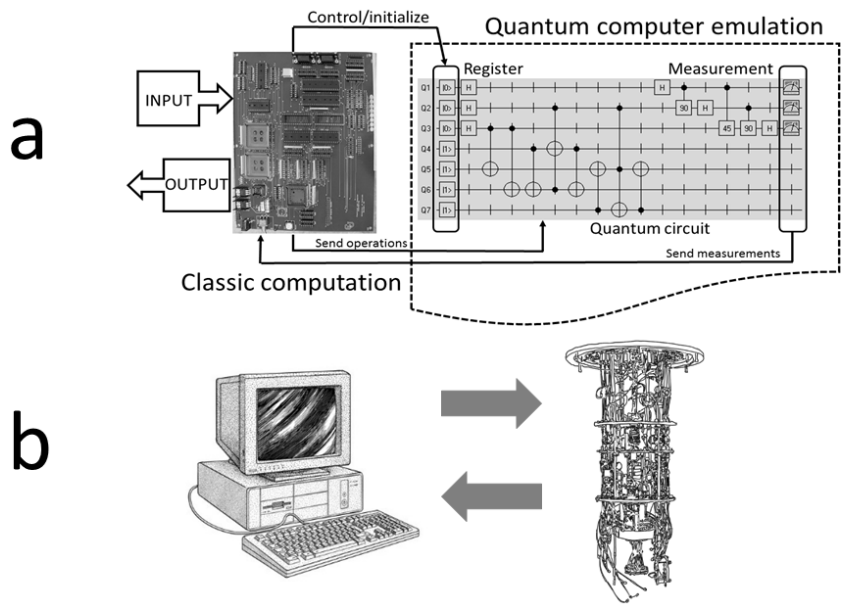

**Figure 4**. (a) "Quantum RAM" (QRAM) model is a hybrid architecture by which a classical computer simultaneously performs classical and quantum operations. According to QRAM model the classical computer performs classical computing, controls the quantum register evolution and send the quantum operations to the quantum computer. The quantum computer, which is really simulated on the classic computer, conducts the initialization of the quantum register state, performs unitary transformations and measurements, sending to the classical computer the output. (b) "Hybrid classical-quantum" computer. In this approach, the main tasks are performed on a real quantum computer, using a classical computer for those operations that cannot be performed on a quantum computer for theoretical or incompatibility reasons.

Assuming we are running the simulation under a QRAM architecture (Figure 4), fitness evaluation (Table IV) is based on $P(t)$ chromosomes and therefore it will require that such calculation takes place on a digital computer.:

**Table IV**. Fitness calculation

| $P(t)$ | $\rightarrow$ | $f(t)$ |
|---|---|---|
| $(x_1\ x_2\ x_3\ ...\ x_j)_1$ | | $f(x_1\ x_2\ x_3\ ...\ x_j)_1$ |
| $(x_1\ x_2\ x_3\ ...\ x_j)_2$ | | $f(x_1\ x_2\ x_3\ ...\ x_j)_2$ |
| , ... , | | , ... , |
| $(x_1\ x_2\ x_3\ ...\ x_j)_i$ | | $f(x_1\ x_2\ x_3\ ...\ x_j)_i$ |

It is important to note that if the fitness were obtained on a quantum computer, then obtaining $P(t)$ from the observation or measurement of $Q(t)$ would cause the destruction of the superposition state leading to the collapse of the wave function. Hence, fitness evaluation is one of the main obstacles in the implementation of QGAs on a quantum computer, e.g. on IBM computer.

A third step consists of updating the population $Q(t)$. For this, we will apply $Q$ gates, e.g. $U(t)$:

$$Q(t + 1) = U(t).Q(t) \quad (5)$$

From a theoretical point of view, population evolution is governed by the Schrödinger equation [13]. Thus, the operator $U(t)$ represents the quantum gates that play the role of the "genetic mechanisms" responsible for introducing genetic variability into the population $Q(t)$. Consequently, the quantum gates are involved in the transformation of the information contained in the chromosomes of one generation for the advantage of the next generation. Since evolution takes place in a complex vector space, i.e. in a Hilbert space, the transformations that take place will be unitary transformations. In a standard QGA, the transformation of the information contained in the chromosomes is the result of applying indistinctly two quantum operators: the rotation or interference operator and the Pauli X-gate (Figure 5). The first of these operators simulates the biological mechanism known as elitism. That is, the purpose of the rotation operator is to modify the chromosomes by bringing them closer to the best evaluated or optimal chromosome. Obviously, the optimal chromosome is the chromosome with the maximum fitness. The second operator, the Pauli X-gate, simulates the genetic mechanism of mutation. Consequently, while rotation promotes the convergence of the population of chromosomes $Q(t)$ to an optimal chromosome, the purpose of mutation is to introduce variability into the population by favoring the exploration of the environment, i.e. the solution space or evolutionary surface. In general, when the Q gates are the rotation operators and the X Pauli gate, then it is a standard QGA or quantum genetic algorithm by default. However, there are quantum genetic algorithms that also include an operator that simulates the recombination or genetic crossover mechanism, either the classical operator of an SGA or some version that claims to be quantum-inspired.

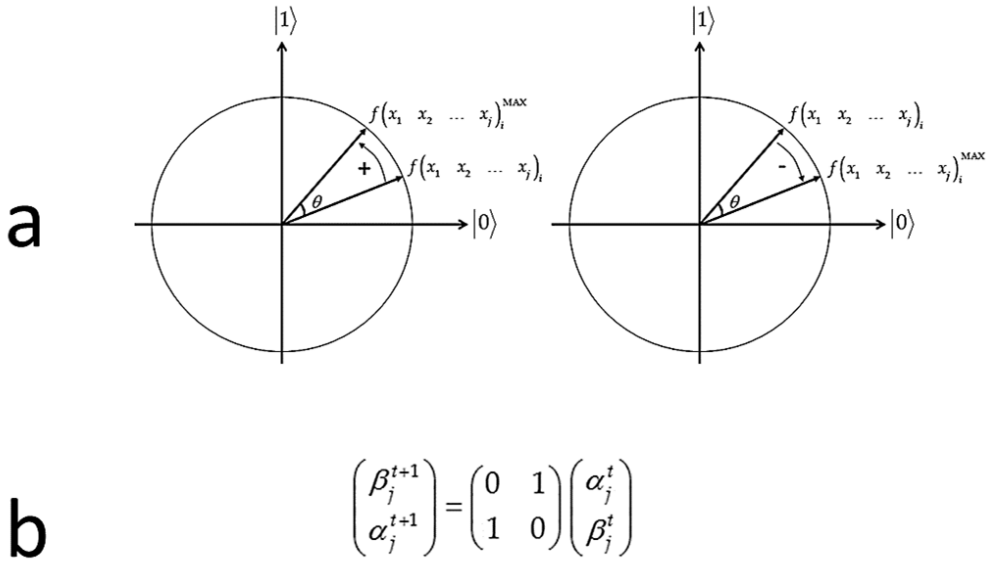

**Figure 5**. (a) Qubit (Interference) Rotation Gate. Applying this operator the evolution of a population is the result of rotations approximating the state of chromosomes to the state of the optimum chromosome in the population. Thus, this gate amplifies or decreases the amplitude of qubits or genes according to the chromosome with maximum fitness. (b) Quantum Mutation (Inversion) Gate. The gate performs an inter-qubit mutation of the *j*th qubit, swapping the amplitudes with the quantum Pauli X gate.

**1.2. Programming a Quantum Genetic Algorithm (QGA)**

In 2015 at the University of Stirling (Scotland, UK) we made a first contact with QGA programming, exploring the possibilities of this kind of algorithms in optimization problems. Since our goal was to understand the fundamentals of quantum computing and to evaluate the usefulness of QGAs, we chose an elementary optimization problem. We selected a benchmark function, the problem being to find the value of *x* that makes *f(x)* maximum in the following function:

$$f(x) = \left| \frac{x-5}{2+\sin(x)} \right| \ , \ 0 \leq x \leq 15 \quad (6)$$

In the selected problem 15 (1011) is the value of *x* for which the function has a maximum value in the studied range. The aim of the experiment was to evaluate the quantum genetic algorithm QGA by studying the convergence to the optimal solution through the performance graph. In 2016 we published a review [14] on quantum genetic algorithms describing through this elementary example the usual types of quantum genetic algorithms and their operators.

Unfortunately, in 2015 the IBM quantum computer [15] was not yet available. For this reason, we performed the simulation experiments on a classical computer with simulated qubits, as well as emulating principles of quantum mechanics such as superposition, interference, etc. We decided to program the QGA in Python language because in this language it is relatively easy to emulate an elementary quantum computer. At that time, and in contrast to the frameworks depicted in Figure 4, this was the most simple and convenient approach for our goals. Indeed, in 2015, the hybrid computing [16] approach (Figure 4b) was not yet available, but it was possible to emulate a quantum computer (Figure 4a) by means of a quantum programming language designed for this specific purpose. One of the earliest quantum programming languages was QCL [17] (the latest available version of QCL is 0.6.7) which emulates a quantum computer assuming the so-called QRAM architecture. An example of QGA coded in QCL language is described by [18] and available at [19].

Once the QGA was implemented in Python 3.4.4 [20] its execution took place on an IBM PC Compatible computer with Windows XP operating system (note that the experiments were conducted in 2015). Thus, we performed simultaneously and without distinction classical operations, e.g. fitness calculation, together with quantum operations, e.g. qubit transformation with quantum gates, register initialization, qubit measurements in $Q(t)$, etc. A detailed explanation of the algorithm steps and the simulation experiment results were described in [14]. Now, what would happen if we tried to run the QGA algorithm we have described on a real quantum computer? QGA simply could not run on a quantum computer like for example IBM. The problem arises from the fact that to date we do not know how to compute the fitness of chromosomes without destroying the superposition state. Note that to obtain the fitness it is necessary, as we have described above, to measure or observe the quantum chromosomes of the population $Q(t)$, destroying their superposition state. That is, the population $Q(t)$ is in a state of superposition and therefore the coherent state of the quantum system is preserved. If we measure in $Q(t)$ the superposition of states is lost, obtaining after the collapse the population $P(t)$. It means that the population $P(t)$ will be in a decoherent state. Being aware of this fact, this limitation does not mean that much of the published literature in which QGAs are used in solving practical real-world problems is useless. In this case, these are algorithms which include some steps inspired by quantum computation and can be run on a classical computer without any problems. However, the execution of these algorithms, and therefore the resolution of the problems for which they were designed, will not be possible to be carried out on a real quantum computer.

## 2. The path to a true QGA: Will we survive crossing the Valley of Death?

One of the most frustrating scenarios in biomedical research conducted by the biotech pharmaceutical industry is popularly known as the "Valley of Death" [21]. This name refers to the great leap that exists between basic research, i.e. when a new drug is discovered with very promising effects against a disease - and its subsequent development and application in the clinical and therapeutic environment. In general, many of these discoveries do not manage to successfully cover all the necessary steps (stages of medications development) from the laboratory to clinical use. After a great investment in work and money, many potentially useful drugs do not succeed in crossing the "Valley of Death" confining the success of the breakthrough to one or more publications in scientific journals.

Currently, a scenario somewhat comparable to drug discovery and development is being played out with the most popular AI algorithms, since the advent of the early quantum computers, many are and have been the attempts to translate these algorithms into their equivalent quantum version. However, many of these attempts have been unsuccessful and have therefore been limited to versions of the original algorithms that, inspired by quantum computing, could not be run on a real quantum computer (Figure 6). How many algorithms will successfully cross the Valley of Death, the cliff that splits classical and quantum computation? As we have pointed out above, one of the main hurdles in implementing a QGA is the computation of the chromosome fitness. The fact of observing or measuring the superposition $Q$(t) causes the loss of the superposition and consequently all the computational "work" done until then would be lost. In order to develop a QGA that could be run on a real quantum computer, e.g. the IBM quantum computer, many years ago a very original idea was put forward: to design a QGA based on Grover's algorithm [22]. This quantum algorithm is able to find in a list of unstructured items an element x with a function $f(x)$ returning $f(x)=1$ when it finds the item x which is the solution of the search. In other words, the item that was intended to be found and which is thus marked. The search takes place from an input in a superposition state and the algorithm amplifies the amplitude of the marked state, which corresponds to the solution.

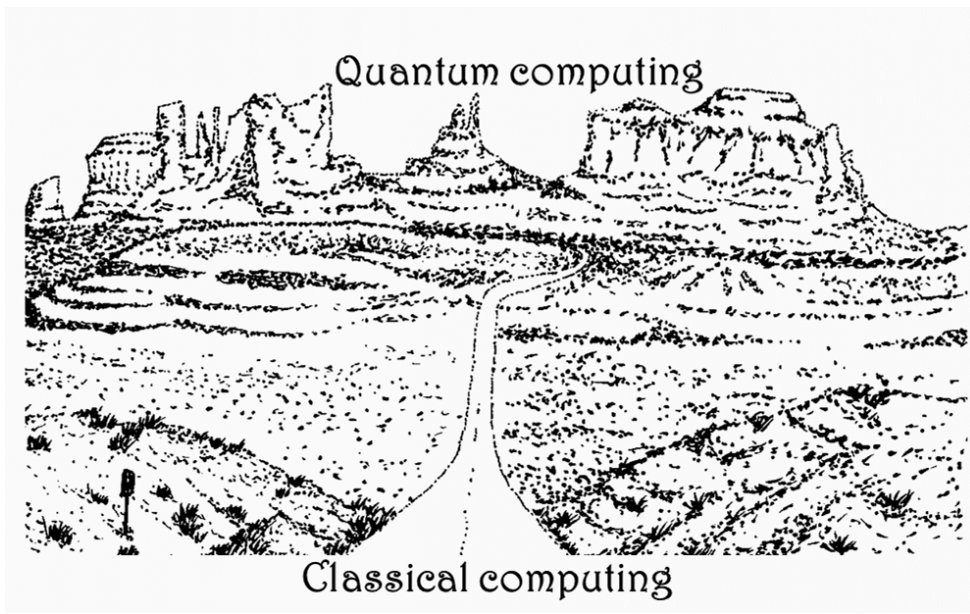

**Figure 6**. "Crossing the Valley of Death". The transition from a classical genetic algorithm to its counterpart quantum version raises challenges that resemble those occurring when, after a new drug has been discovered, it is intended to be launched on the market and applied at the clinical level.

In 2006 Udrescu et al. [23] presented a QGA based on the embedding of an F-gate in a quantum circuit implementing a Grover algorithm (Figure 7). In this algorithm the F-gate (Figure 8) could evaluate the fitness of chromosomes without destroying the superposition. This idea was also cited by [24] Kumar and Goswani in 2013. In Table V we show the main steps of this algorithm, which was named as RQGA (an acronym for Reduced Quantum Genetic Algorithm).

**Table V.** Main steps of a RQGA.

| Step | Quantum Computing |
|------|-------------------|
| 1 | Initialize a superposition of all possible chromosomes |
| 2 | Evaluates fitness with operator $F$ |
| 3 | Ask to the oracle $O$ |
| 4 | Apply Grover's diffusion operator $G$ |
| 5 | Make a measure |

Indeed, the possibility of using Grover's algorithm as a search protocol for the optimal solution was an idea that emerged in these researchers before the first quantum computers were available. At first sight, the idea of using Grover's algorithm as an essential component of a QGA was really interesting since it opened the possibility of simulating Darwinian evolution in a quantum computer, either for theoretical studies or practical purposes through the design of QGAs. Consequently, using the protocol of Table V it would be possible to implement the quantum version of an optimization algorithm with an evolutionary structure. In addition, using Grover's algorithm to implement a QGA leads to some interesting ideas. For example, the genetic search strategy based on a gradual, tree-based exploration characteristic of Darwinian evolution is replaced by a search based on Grover's algorithm which implies a smaller number of steps during the search. Thus, assuming that the Darwinian search is a tree search, the complexity will be $O(N)$, where $N$ is the number of nodes, individuals or possible solutions. Now, when the search is via Grover algorithm then the evolutionary search is reduced to $O(\sqrt{N})$.

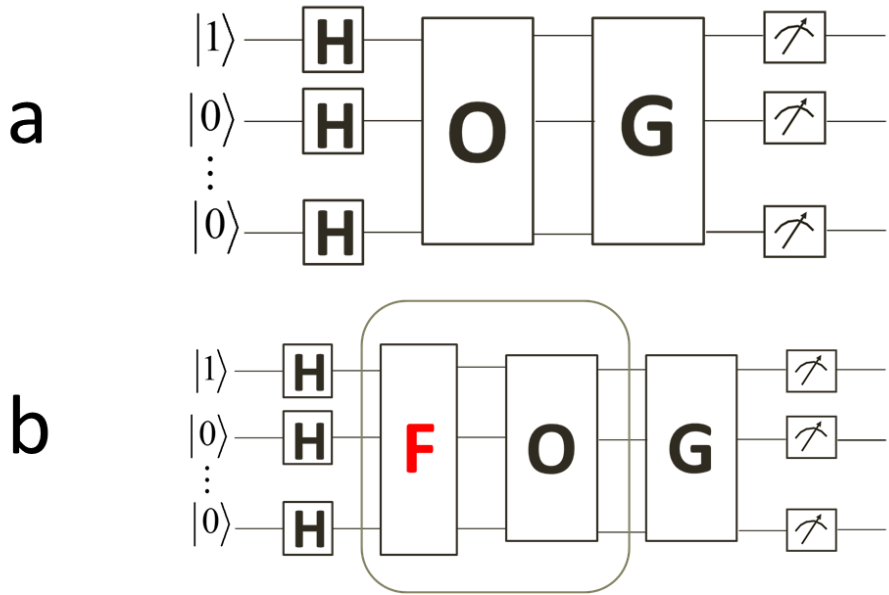

**Figure 7**. (a) Grover and (b) RQGA circuits.

Now, similar to a QGA, in an RQGA the population $Q(t)$ is also represented by a superposition, with all individuals included in a single quantum register. In this regard, then, what is the novelty introduced by an RQGA? The new feature is that in the RQGA the quantum register of individuals or chromosomes is correlated with a quantum register of the fitness values of the chromosomes. In other words, the fitness of the chromosomes of the population $Q(t)$ would be evaluated by a quantum fitness gate F [22, 23] but F does not destroy the superposition:

$$F|\psi\rangle_i|0\rangle = \frac{1}{\sqrt{n}}\sum_{i=0}^{2^n-1}|x_i\rangle \otimes |fitness_x\rangle_i \quad (7)$$

Therefore, the interesting feature of this fitness operator is that by using F (Figure 8) only once we could obtain all the fitness values of the chromosomes of the population *P*(*t*), without *sacrificing* the superposition *Q*(t).

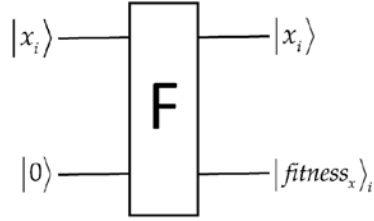

**Figure 8**. Quantum fitness operator

In view of this striking feature of the quantum genetic algorithm RQGA, in the study on QGAs conducted in 2015 [14] we included the analysis of RQGA. The aim of this analysis was to figure out how RQGA could be applied to solve a real case, using for this purpose the elementary optimization problem (6) described above.

Next, we will describe step by step how to apply the RQGA algorithm to solve the referred optimization problem (see program 1 in the Appendix).

## 2.1. Running the RQGA algorithm with a classical computer

First, in order to find the maximum value of the function (6) with RQGA, we will prepare all the individuals of the superposition that defines the population *Q*(t):

$$|\psi\rangle_i = \frac{1}{\sqrt{n}} \sum_{i=0}^{2^n-1} |x_i\rangle = \frac{1}{4} \begin{pmatrix} |0000\rangle \\ +|0001\rangle \\ +|0010\rangle \\ +|0011\rangle \\ +|0100\rangle \\ +|0101\rangle \\ +|0110\rangle \\ +|0111\rangle \\ +|1000\rangle \\ +|1001\rangle \\ +|1010\rangle \\ +|1011\rangle \\ +|1100\rangle \\ +|1101\rangle \\ +|1110\rangle \\ +|1111\rangle \end{pmatrix} \quad (8)$$

Next, if we apply the quantum fitness operator to the optimization problem (6) then we will have the following quantum system:

$$F|\psi\rangle_i|0000000000\rangle = \frac{1}{\sqrt{n}}\sum_{i=0}^{2^n-1}|x_i\rangle\otimes|fitness_x\rangle_i = \frac{1}{4}\begin{pmatrix}|0000\rangle\\+|0001\rangle\\+|0010\rangle\\+|0011\rangle\\+|0100\rangle\\+|0101\rangle\\+|0110\rangle\\+|0111\rangle\\+|1000\rangle\\+|1001\rangle\\+|1010\rangle\\+|1011\rangle\\+|1100\rangle\\+|1101\rangle\\+|1110\rangle\\+|1111\rangle\end{pmatrix}\otimes|0000000000\rangle \quad (9)$$

As mentioned by [22, 23] we assume that operator F is able to compute the fitness values of the individuals in the population $Q(t)$. We calculated the fitness of the chromosomes as shown between lines 58-85 of program 1 (Appendix). Once the fitness of the chromosomes has been calculated, we will write (9) as follows:

$$|\psi\rangle_i = \frac{1}{\sqrt{n}}\sum_{i=0}^{2^n-1}|x_i\rangle\otimes|fitness_x\rangle_i = \frac{1}{4}\begin{pmatrix}|0000\rangle\otimes|0011111010\rangle\\+|0001\rangle\otimes|0010001100\rangle\\+|0010\rangle\otimes|0001100111\rangle\\+|0011\rangle\otimes|0001011101\rangle\\+|0100\rangle\otimes|0001010000\rangle\\+|0101\rangle\otimes|0000000000\rangle\\+|0110\rangle\otimes|0000111010\rangle\\+|0111\rangle\otimes|0001001011\rangle\\+|1000\rangle\otimes|0001100100\rangle\\+|1001\rangle\otimes|0010100101\rangle\\+|1010\rangle\otimes|0101010111\rangle\\+|1011\rangle\otimes|1001010111\rangle\\+|1100\rangle\otimes|0111011110\rangle\\+|1101\rangle\otimes|0101001010\rangle\\+|1110\rangle\otimes|0100101100\rangle\\+|1111\rangle\otimes|0101111001\rangle\end{pmatrix} \quad (10)$$

At this point in the present experiment, what is the main pitfall of the algorithm? Although the replacement of the Darwinian search by another one based on Grover's algorithm is quite original, none of the papers [22, 23, 24] presenting the RQGA offers indications or guidelines on how to calculate fitness values in practice. Since nothing is said about how to design the operator F we were advocated to calculate fitness from the information contained in the chromosomes, i.e., from the population $P(t)$. Therefore, in the RQGA genetic algorithm we will also need to collapse the quantum version of the population $Q(t)$ as long as no details are given about F implementation. The result will be the decoherent population $P(t)$ from which we calculate the fitness of the chromosomes. Next, we will search for that individual or chromosome with maximum fitness $\left| fitness_x \right\rangle_i^{max}$. In summary, and after all this interesting theoretical background but without any indication of how to put it into practice, all that remains is to obtain the fitness values and their maximum value using a simple algorithm that is run on a classic computer.

As shown in (10) the fitness values were converted to their corresponding binary value. Although at present the issue of fitness calculation remains unsolved we will now outline the remaining steps of the RQGA algorithm. After calculating the fitness of the chromosomes, we proceed with the following steps.

Second, similar to a standard QGA, we initialize the superposition of all possible chromosomes, i.e., we conduct the initialization of the quantum register (1). As a result, we obtain a superposition state (program 1, lines 125-129) with probabilities:

$$\begin{pmatrix} 0.25 \\ 0.25 \\ 0.25 \\ 0.25 \\ 0.25 \\ 0.25 \\ 0.25 \\ 0.25 \\ 0.25 \\ 0.25 \\ 0.25 \\ 0.25 \\ 0.25 \\ 0.25 \\ 0.25 \\ 0.25 \end{pmatrix} \quad (11)$$

corresponding to the sixteen possible chromosome states represented by:

$$|0\rangle = \begin{pmatrix}1\\0\\0\\0\\0\\0\\0\\0\\0\\0\\0\\0\\0\\0\\0\\0\end{pmatrix}, |1\rangle = \begin{pmatrix}0\\1\\0\\0\\0\\0\\0\\0\\0\\0\\0\\0\\0\\0\\0\\0\end{pmatrix}, |2\rangle = \begin{pmatrix}0\\0\\1\\0\\0\\0\\0\\0\\0\\0\\0\\0\\0\\0\\0\\0\end{pmatrix}, \ldots |11\rangle = \begin{pmatrix}0\\0\\0\\0\\0\\0\\0\\0\\0\\0\\0\\1\\0\\0\\0\\0\end{pmatrix}, \ldots |15\rangle = \begin{pmatrix}0\\0\\0\\0\\0\\0\\0\\0\\0\\0\\0\\0\\0\\0\\0\\1\end{pmatrix} \quad (12)$$

that is $|0\rangle = |0000\rangle_0$, $|1\rangle = |0001\rangle_1$, $|2\rangle = |0010\rangle_2, \ldots, |11\rangle = |1011\rangle_{11}, \ldots, |15\rangle = |1111\rangle_{15}$.

Thirdly, and according to Grover's algorithm the oracle $O$ will mark the individual or chromosome $|\psi\rangle_i$ with the maximum fitness value. Therefore, once the RQGA algorithm queries an oracle $O$, i.e. $O|\psi\rangle^{Q(t)}$, $O$ answers by marking the optimal chromosome with an eigenvalue equal to $(-1)^{f(x)}$:

$$O|\psi\rangle^{Q(t)} = (-1)^{f(x)} |\psi\rangle^{Q(t)} \quad (13)$$

such that:

$$f(x) = \begin{cases} 1, & if \ |fitness_x\rangle_i = |fitness_x\rangle_i^{max} \\ 0, & otherwise \end{cases} \quad (14)$$

Consequently, and once oracle $O$ has marked with -1 the individual or chromosome with maximum fitness value, we will obtain the matrix (program 1, line 134):

$$\begin{pmatrix}
1 & 0 & 0 & 0 & 0 & 0 & 0 & 0 & 0 & 0 & 0 & 0 & 0 & 0 & 0 & 0 \\
0 & 1 & 0 & 0 & 0 & 0 & 0 & 0 & 0 & 0 & 0 & 0 & 0 & 0 & 0 & 0 \\
0 & 0 & 1 & 0 & 0 & 0 & 0 & 0 & 0 & 0 & 0 & 0 & 0 & 0 & 0 & 0 \\
0 & 0 & 0 & 1 & 0 & 0 & 0 & 0 & 0 & 0 & 0 & 0 & 0 & 0 & 0 & 0 \\
0 & 0 & 0 & 0 & 1 & 0 & 0 & 0 & 0 & 0 & 0 & 0 & 0 & 0 & 0 & 0 \\
0 & 0 & 0 & 0 & 0 & 1 & 0 & 0 & 0 & 0 & 0 & 0 & 0 & 0 & 0 & 0 \\
0 & 0 & 0 & 0 & 0 & 0 & 1 & 0 & 0 & 0 & 0 & 0 & 0 & 0 & 0 & 0 \\
0 & 0 & 0 & 0 & 0 & 0 & 0 & 1 & 0 & 0 & 0 & 0 & 0 & 0 & 0 & 0 \\
0 & 0 & 0 & 0 & 0 & 0 & 0 & 0 & 1 & 0 & 0 & 0 & 0 & 0 & 0 & 0 \\
0 & 0 & 0 & 0 & 0 & 0 & 0 & 0 & 0 & 1 & 0 & 0 & 0 & 0 & 0 & 0 \\
0 & 0 & 0 & 0 & 0 & 0 & 0 & 0 & 0 & 0 & 1 & 0 & 0 & 0 & 0 & 0 \\
0 & 0 & 0 & 0 & 0 & 0 & 0 & 0 & 0 & 0 & 0 & -1 & 0 & 0 & 0 & 0 \\
0 & 0 & 0 & 0 & 0 & 0 & 0 & 0 & 0 & 0 & 0 & 0 & 1 & 0 & 0 & 0 \\
0 & 0 & 0 & 0 & 0 & 0 & 0 & 0 & 0 & 0 & 0 & 0 & 0 & 1 & 0 & 0 \\
0 & 0 & 0 & 0 & 0 & 0 & 0 & 0 & 0 & 0 & 0 & 0 & 0 & 0 & 1 & 0 \\
0 & 0 & 0 & 0 & 0 & 0 & 0 & 0 & 0 & 0 & 0 & 0 & 0 & 0 & 0 & 1
\end{pmatrix} \quad (15)$$

The population $Q(t)$ that is in the superposition state in accordance with the RQGA algorithm will have marked the optimal solution, i.e. the chromosome state eleven (program 1, lines 136-137):

$$\begin{pmatrix} 0.25 \\ 0.25 \\ 0.25 \\ 0.25 \\ 0.25 \\ 0.25 \\ 0.25 \\ 0.25 \\ 0.25 \\ 0.25 \\ 0.25 \\ -0.25 \\ 0.25 \\ 0.25 \\ 0.25 \\ 0.25 \end{pmatrix} \quad (16)$$

Next, we apply Grover's diffusion operator (program 1, lines 109-112, 139-140) obtaining:

$$\begin{pmatrix} 0.1875 \\ 0.1875 \\ 0.1875 \\ 0.1875 \\ 0.1875 \\ 0.1875 \\ 0.1875 \\ 0.1875 \\ 0.1875 \\ 0.1875 \\ 0.1875 \\ 0.6875 \\ 0.1875 \\ 0.1875 \\ 0.1875 \\ 0.1875 \end{pmatrix} \quad (17)$$

Finally, if we repeat the above steps several times and finally we measure, we will obtain the value of *x* for which *f(x)* is maximum. The solution is given by state $|11\rangle$ which corresponds to the individual or chromosome $|1011\rangle_{11}$.

**2.2. Running RQGA on IBM quantum computer.**

The review [14] published in 2016 on quantum genetic algorithms covered the usual types of QGA, including the RQGA we have just discussed. In 2002 a web page [25] discloses the example of the function (6) described above to explain the RQGA algorithm, programming the algorithm in Qiskit language [26] for the IBM quantum computer (see program 2 in the Appendix). The optimization problem that we have solved with RQGA is also applied to other functions in [26] as well as in another study published later by [27]. A remarkable feature is that in these two surveys [26, 27] the optimization problem is also restricted to the use of four qubits, and therefore to sixteen possible states shown in (12).

Also in 2022 Ardelean and Udrescu [28] describe the RQGA and its potential applications are illustrated through the graph coloring problem, including the code in Qiskit language [29]. Likewise, the RQGA algorithm is implemented at the circuit level [30] using Python and Qiskit. Finally, in 2023 [31] these two researchers introduce an improved version of the RQGA algorithm, which they term HQAGO (Hybrid Quantum Algorithm with Genetic Optimization).

Nevertheless, to date, the problem of calculating the fitness of chromosomes without breaking the *Q(t)* superposition remains unsolved. Evidence of this statement is the fact that the input of program 2 (Appendix) is the output (15) of oracle *O*. This is the protocol followed in [26, 27]. In other words, the fitness of the chromosomes has been previously calculated "outside a quantum computer" in a similar way as we did with program 1 (Appendix) written in Python on a classical computer. Likewise, the RQGA algorithm implementation in Qiskit [29] should be run on a real IBM quantum computer in order to verify whether RQGA is indeed a true quantum genetic algorithm. The genetic algorithm RQGA was proposed as a very creative solution to the difficulties arising when implementing the genetic operators of a classical genetic algorithm in a quantum computer [32]. The adoption of Grover's algorithm as a selection operator was an interesting step forward because it motivated further research in the design of new QGAs. However, to date the RQGA algorithm can only be run on a classical computer with QRAM or hybrid architecture, or alternatively by resorting to the services of the IBM quantum simulator.

In summary, the RQGA genetic algorithm is basically the application of Grover's algorithm to a list of chromosomes or solutions for which their fitness values have been previously calculated. Obviously, Grover's algorithm (program 2) - one of the most popular quantum algorithms [22] - can be implemented in Qiskit language and be executed on a real IBM quantum computer.

## 3. Conclusion and Discussion

In this paper we have described the challenges arising when, inspired by the steps of an SGA algorithm, we try to reach its quantum QGA counterpart. Obviously, the challenges we have described in designing a real QGA are illustrative of the kind of issues and difficulties arising when attempting to develop a quantum version of an AI algorithm that was originally designed for a classical computer.

In our opinion, these shortcomings may have their origin in the mistaken inclination to establish a parallelism and even an equivalence between classical or von Neumann computers and quantum computers. In general, we adopt an analogy between bit and qubit, logic gates (AND, OR, etc.) and quantum gates (Pauli-X, Hadamard, Toffoli, etc.), CPU and a quantum processor, RAM memory and a quantum register, etc. when the similarity between these elements is more symbolic than physical. Adopting perhaps unconsciously this misconception leads inexorably to an attempt to translate one to one the steps or tasks of a classical algorithm to its quantum counterpart. Consequently, we often forget that a quantum computer shares hardly any features with a classical one and, moreover, it cannot perform many of the latter's functions. Thus, a pure quantum computer cannot perform [33] conditional tasks (if then), loops (for to), function calls, floating point numbers, etc. Regarding the latter, many researchers learned to program on the basis of the so-called structured program or Böhm-Jacopini theorem. According to this theorem, any function computable by a Turing machine, i.e. a classical computer, can be implemented in a programming language that combines three control structures: sequential, conditional and loop or iteration structures. However, a quantum computer processes the information in batches, i.e. sequentially, and neither loops nor conditional expressions can be executed. Therefore, this fact could be another source of many of the problems we face when implementing the quantum version of an AI algorithm.

The conclusion is therefore that the lack of enough resemblance between a digital computer and a quantum computer, and the absence of programming structures that are present in a digital computer, should lead us to change the a priori perception we have about what a quantum computer really is. Also, the enthusiasm of the scientific community towards these computers, which are still at a young stage of their development, has led in many instances to a certain disappointment and frustration on seeing the current limitations of today's quantum computers [34]. For this reason, and those we have mentioned above, the use of hybrid computing is in many cases the best and most practical solution for the implementation of many AI algorithms. For instance, how to implement the learning rule of a perceptron neural network in which the weights of the network are adjusted iteratively? In a machine learning model, if a decision needs to be made, how to implement the if-then structure? Indeed, resorting to hybrid computing paradigm or running on IBMQ QASM cloud quantum simulator can be a good solution. For instance, the artificial neural network [35] or the quantum-inspired evolutionary algorithm [36] are two good examples of quantum inspired algorithms running on IBMQ QASM cloud quantum simulator. Now, it is also possible to adopt a hybrid computation by interacting the processor of a classical computer with the family of quantum processors provided by the IBM Q Experience initiative. This last methodology was applied by [37] to design a hybrid quantum genetic algorithm (HQGA) for solving the 0-1 knapsack problem. Likewise, [38] introduced an HQGA which performance was studied with different discretized benchmark functions.

Obviously, there are also 'true' quantum algorithms, i.e. machine learning algorithms that can be run on a quantum computer and therefore are part of the so-called Quantum Artificial Intelligence [39]. For instance, [40] successfully tested on a quantum processor an optical character recognition system based on an experimental implementation of the quantum version of the SVM (Support Vector Machine) algorithm.

In this paper, using genetic algorithms as a case study, we have shown the difficulties that can arise when an algorithm that was developed for digital computers is translated to its corresponding quantum version. Indeed, we are still in the early days of quantum computers, but it is obvious that if we want to have a successful development of Quantum Artificial Intelligence we will have to address a number of issues. The solution of these issues will imply that we change our way of thinking and the way we use this new class of computers.

## 4. Appendix

**Program 1** (see [41])

```python
import math
import numpy as np

n=4;
fitness = np.empty([2**n])
cutoff=99;

def bin2dec(string_num):
    return str(int(string_num, 2))

def dec2vec(dec,n):
    vec=np.zeros((2**n,1))
    vec[dec,0]=1;
    return vec

def psi(string_num):
    dec=bin2dec(string_num)
    return dec2vec(dec,len(string_num))

def hadamard(n):
    r2=math.sqrt(2.0)
    H1=np.array([[1/r2,1/r2],[1/r2,-1/r2]])
    if n==1:
        H=H1
    else:
        H=1;
        i=1;
        for i in range(1,n+1):
            H=np.kron(H,H1)
    return H

def bin(i):
    if i == 0:
        return "0"
    s = ''
    while i:
        if i & 1 == 1:
            s = "1" + s
        else:
            s = "0" + s
        i >>= 1
    return s
```

```python
def Fitness_evaluation():
    for i in range(0,2**n):
        fitness[i]=0

    ##########################################################
    # Define your problem in this section. For instance:     #
    # f(x)=abs(x-5/2+sin(x)) , 0<=x<=15                      #
    # maximum value at x=11 (x binary is 1011)               #
    ##########################################################
    for i in range(0,16):
        x=int(bin(i),2)
        # replaces the value of x in the function f(x)
        y=np.fabs((x-5)/(2+np.sin(x)))
        # the fitness value is calculated below:
        # (Note that in this example is multiplied
        # by a scale value, e.g. 100)
        fitness[i]=y*100

    ##########################################################
    # Best chromosome selection
    the_best_chrom=0;
    fitness_max=fitness[1];
    for i in range(0,16):
        if fitness[i]>=fitness_max:
            fitness_max=fitness[i]
            the_best_chrom=i
        cutoff=the_best_chrom
    return cutoff

##########################################################
# ORACLE:UNITARY F-CONDITIONAL INVERTER FOR N BITS INPUT#
##########################################################
def U_Oracle(n):
    zero_mat=np.zeros((2**n,2**n))
    i=0;
    for i in range(0,2**n):
    ############################
    # Define here your oracle #
    ############################
        if i==Fitness_evaluation():
            O=1
        else:
            O=0
    # Inverter
        zero_mat[i,i]=(-1)**O
    return zero_mat
```

```python
#########################################################
# GROVER'S DIFFUSION OPERATOR                           #
#########################################################
# Inversion about average
def ia(n):
    ia_mat=2*np.ones(2**n)/(2**n)
    ia_mat=ia_mat-np.identity(2**n)
    return ia_mat

#########################################################
# GROVER'S MAXIMUM ITERATIONS                           #
#########################################################
def maxiter(n):
    max_iter=(np.pi/4)*math.sqrt(2**n)
    return max_iter

#########################################################
# REDUCED QUANTUM GENETIC ALGORITHM                     #
#########################################################
def RQGA(n, string_num):
    psi_=psi(string_num)
    H=hadamard(n)
    psi_=np.dot(H,psi_)
    print(psi_)
    print()
    iter=np.trunc(maxiter(n))
    iter=int(round(iter))
    for i in range (1,iter):
        U_O=U_Oracle(n)
        print(U_O)
        print()
        psi_=np.dot(U_O,psi_)
        print(psi_)
        print()
        D=ia(n)
        psi_=np.dot(D,psi_)
    print(psi_)

#########################################################
# MAIN PROGRAM                                          #
#########################################################
print("REDUCED QUANTUM GENETIC ALGORITHM")
input("Press Enter to continue...")
RQGA(4,'0000');
```

**Program 2** (see [26])

```
In [1]:  from qiskit import *
         from qiskit.tools.visualization import plot_histogram
         import math
         import numpy as np
         %matplotlib inline
```

```python
In [2]: from qiskit import *
        from qiskit.tools.visualization import plot_histogram
        %matplotlib inline

        n = 4

        #Init population
        circuit = QuantumCircuit(n)#,2)
        circuit.h(range(n))
        rand_theta = np.random.uniform(0, np.pi/2, 4)
        print(rand_theta)
        for i in range(n):
            circuit.rz(rand_theta[i], i)

        circuit.draw(output = "mpl" ,style = {'backgroundcolor':'gray'})
```

```
[0.97576923 0.58408783 1.36733896 0.82551891]
```

Out[2]:
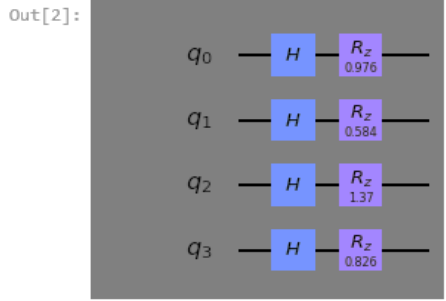

```python
In [3]: from qiskit.quantum_info.operators import Operator
        ############################## fitness function oracle for the problem mensioned below #####
        ##########################################################
        #                                                        #
        # Problem: optimize a function f(x)=abs(x-5/2+sin(x))    #
        # values in the range 0<=x<=15. Within this range f(x)   #
        # has a maximum value at x=11 (binary is equal to 1011)  #
        ##########################################################
        ################################################################################
        f = [
            [1,0,0,0,0,0,0,0,0,0,0,0,0,0,0,0],
            [0,1,0,0,0,0,0,0,0,0,0,0,0,0,0,0],
            [0,0,1,0,0,0,0,0,0,0,0,0,0,0,0,0],
            [0,0,0,1,0,0,0,0,0,0,0,0,0,0,0,0],
            [0,0,0,0,1,0,0,0,0,0,0,0,0,0,0,0],
            [0,0,0,0,0,1,0,0,0,0,0,0,0,0,0,0],
            [0,0,0,0,0,0,1,0,0,0,0,0,0,0,0,0],
            [0,0,0,0,0,0,0,1,0,0,0,0,0,0,0,0],
            [0,0,0,0,0,0,0,0,1,0,0,0,0,0,0,0],
            [0,0,0,0,0,0,0,0,0,1,0,0,0,0,0,0],
            [0,0,0,0,0,0,0,0,0,0,1,0,0,0,0,0],
            [0,0,0,0,0,0,0,0,0,0,0,-1,0,0,0,0],
            [0,0,0,0,0,0,0,0,0,0,0,0,1,0,0,0],
            [0,0,0,0,0,0,0,0,0,0,0,0,0,1,0,0],
            [0,0,0,0,0,0,0,0,0,0,0,0,0,0,1,0],
            [0,0,0,0,0,0,0,0,0,0,0,0,0,0,0,1]
        ]
        circuit.unitary(f,range(n), label='F')
        circuit.draw(output = "mpl" ,style = {'backgroundcolor':'gray'})
```

Out[3]:
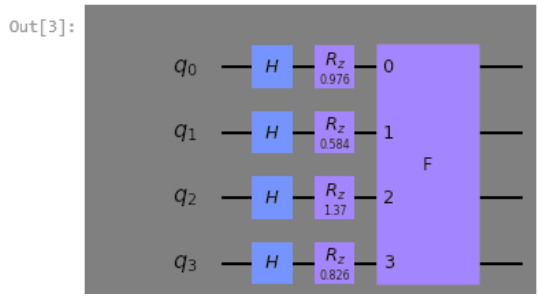

```python
In [5]: #Gr_Inv_Mat
        def ia(n):
            ia_mat=2*np.ones(2**n)/(2**n)
            ia_mat=ia_mat-np.identity(2**n)
            return ia_mat
        GIM = ia(n)
        circuit.unitary(GIM,range(n), label='GIM')

        #GN
        gn = int((np.pi/4)*(np.sqrt(2**n)))
        #print(gn)
        for i in range(gn):
            circuit.unitary(f,range(n), label='F')
            circuit.unitary(GIM,range(n), label='GIM')
        #circuit.measure(range(2),range(2))
        circuit.draw(output = "mpl" ,style = {'backgroundcolor':'gray'})
```

Out[5]:

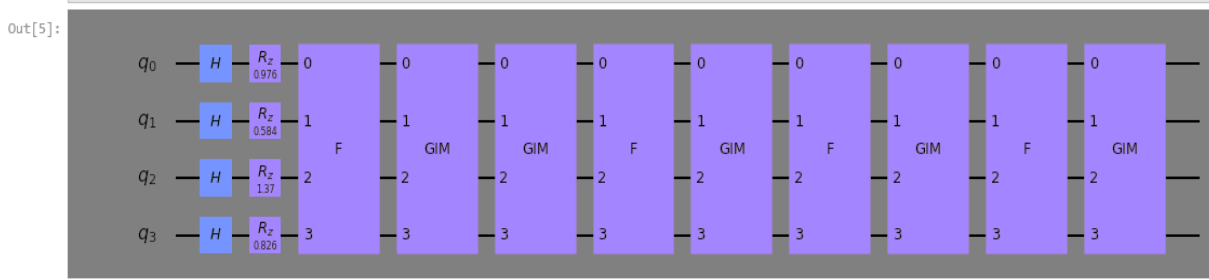

```python
In [6]: circuit.measure_all()
```

```python
In [7]: #backend
        from collections import Counter

        sim = Aer.get_backend('qasm_simulator')
        result = Counter(execute(circuit,backend=sim,shots=100).result().get_counts())
        print(result)
        #creating histogram
        from qiskit.tools.visualization import plot_histogram
        plot_histogram(result)
```

Counter({'1011': 34, '0000': 13, '0010': 9, '1111': 9, '0001': 8, '0111': 7, '1000': 5, '1101': 3, '1010': 3, '0100': 3, '0101': 3, '1110': 2, '0011': 1})

Out[7]:

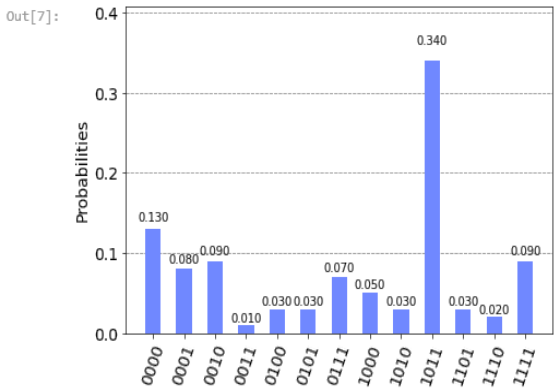

```python
In [15]: def bin2dec(string_num):
             return int(string_num, 2)
```

```python
In [9]: bin2dec(result.most_common()[0][0])
```

Out[9]: 11

**Acknowledgments**

I thank Professor Biswajit Basu for his invitation to give a conference in the Workshop on Quantum Algorithms, Machine Learning and Control (QUAMCON 2023) organized by Trinity College Dublin School of Engineering. This paper was presented at the aforementioned Workshop.

**Funding:** I would like to express my thanks to the Trinity College Dublin School of Engineering for the funding received which enabled me to attend the Workshop held on June 22-23, 2023.